\documentclass{article}

    \usepackage[preprint,nonatbib]{neurips_2024}

\usepackage[utf8]{inputenc} % allow utf-8 input
\usepackage[T1]{fontenc}    % use 8-bit T1 fonts
\usepackage{hyperref}       % hyperlinks
\usepackage{url}            % simple URL typesetting
\usepackage{booktabs}%       % professional-quality tables
\usepackage{amsfonts}       % blackboard math symbols
\usepackage{nicefrac}       % compact symbols for 1/2, etc.
\usepackage{microtype}      % microtypography
\usepackage{xcolor}         % colors

\usepackage{multirow}%
\usepackage{amsmath,amssymb,amsfonts}%
\usepackage{mathrsfs}%
\usepackage[title]{appendix}%
\usepackage{textcomp}%
\usepackage{graphicx}%
\usepackage{manyfoot}%
\usepackage{tabularx}
\usepackage{hhline}
\usepackage{pifont}% http://ctan.org/pkg/pifont
\usepackage{array}
\usepackage{siunitx}

\usepackage{mwe} % to get dummy images
\usepackage{bbm}
\usepackage{adjustbox}
\usepackage{csquotes}

\newcommand{\etal}{\textit{et al}.\xspace}
\newcommand{\ie}{\textit{i}.\textit{e}.\xspace}
\newcommand{\eg}{\textit{e}.\textit{g}.\xspace}

\definecolor{jimcolor}{RGB}{83, 158, 198}
\definecolor{blue_munsell}{rgb}{0.36, 0.54, 0.66}
\definecolor{blue-violet}{rgb}{0.54, 0.17, 0.89}
\definecolor{byzantine}{rgb}{0.74, 0.2, 0.64}
\definecolor{caputmortuum}{rgb}{0.35, 0.15, 0.13}
\definecolor{alizarin}{rgb}{0.82, 0.1, 0.26}
\definecolor{light_grey}{rgb}{0.6, 0.6, 0.6}

\makeatletter
\def\adl@drawiv#1#2#3{%
        \hskip.5\tabcolsep
        \xleaders#3{#2.5\@tempdimb #1{1}#2.5\@tempdimb}%
                #2\z@ plus1fil minus1fil\relax
        \hskip.5\tabcolsep}
\newcommand{\cdashlinelr}[1]{%
  \noalign{\vskip\aboverulesep
           \global\let\@dashdrawstore\adl@draw
           \global\let\adl@draw\adl@drawiv}
  \cdashline{#1}
  \noalign{\global\let\adl@draw\@dashdrawstore
           \vskip\belowrulesep}}
\makeatother

\usepackage[backend=bibtex]{biblatex} % Uncomment for arxiv!!!
\title{Random Token Fusion for Multi-View Medical Diagnosis}

\author{\noindent
\textbf{Jingyu Guo} $^{1,2}$
\thanks{Corresponding author <jingyug@kth.se>} 
\vspace{2mm} \hspace{4mm}
\textbf{Christos Matsoukas} $^{1,2}$ \hspace{4mm}
\textbf{Fredrik Strand} $^{3,4}$ \hspace{4mm}
\textbf{Kevin Smith} $^{1,2}$\\
\normalfont{\noindent
$^{1}$ KTH Royal Institute of Technology, Stockholm, Sweden} \\
$^{2}$ Science for Life Laboratory, Stockholm, Sweden \\
$^{3}$ Karolinska Institutet, Stockholm, Sweden \\
$^{4}$ Karolinska University Hospital, Stockholm, Sweden
}

\begin{document}

\maketitle

% % % % % % % % % 

\begin{abstract}

In multi-view medical diagnosis, deep learning-based models often fuse information from different imaging perspectives to improve diagnostic performance. 
However, existing approaches are prone to overfitting and rely heavily on view-specific features, which can lead to trivial solutions.
In this work, we introduce Random Token Fusion (RTF), a novel technique designed to enhance multi-view medical image analysis using vision transformers. 
By integrating randomness into the feature fusion process during training, RTF addresses the issue of overfitting and enhances the robustness and accuracy of diagnostic models without incurring any additional cost at inference.
We validate our approach on standard mammography and chest X-ray benchmark datasets. 
Through extensive experiments, we demonstrate that RTF consistently improves the performance of existing fusion methods, paving the way for a new generation of multi-view medical foundation models.

\end{abstract}

\section{Introduction}
\label{intro}

Physicians routinely employ multi-view analysis in diagnostic procedures. 
Images gathered at various angles can unveil details that may be obscured in a single view, enhancing the precision of the diagnosis.
It has been shown in clinical trials that a large percentage of breast cancers can only be detected when both craniocaudal (CC) and mediolateral oblique (MLO) views are analyzed \cite{HACKSHAW2000454}.
Similarly, the frontal and lateral views of chest X-rays can provide unique information that is valuable to the accurate diagnosis of diseases \cite{feigin_2010, pmlr-v121-hashir20a, raoof_feigin_sung_raoof_irugulpati_rosenow_2012, Irvin_2019}. 
This is because complementary information from different views helps physicians tackle the challenges posed by superimposed tissues and complex anatomy. 
Each additional view provides context and offers unique insights, enriching the overall understanding of a patient's condition \cite{Irvin_2019}.

Given the clear benefits of multi-view analysis in clinical practice, it stands to reason that foundation models for medical image analysis could similarly improve their diagnostic accuracy by integrating information from multiple views.
Information from different views of anatomical structures may reduce ambiguity, better explain spatial relationships, and provide additional context. 
Although only a handful of works have addressed the topic of multi-view fusion in neural networks, evidence suggests that there is a tangible benefit from fusing information from multiple views 
\cite{Chen2022TransformersIB, black2024multi, Wu2020ImprovingTA, Zhu2021MVC, 10.1007/978-3-030-87199-4_10}.

\begin{figure}[t]
    \centering
    \includegraphics[width=\linewidth]{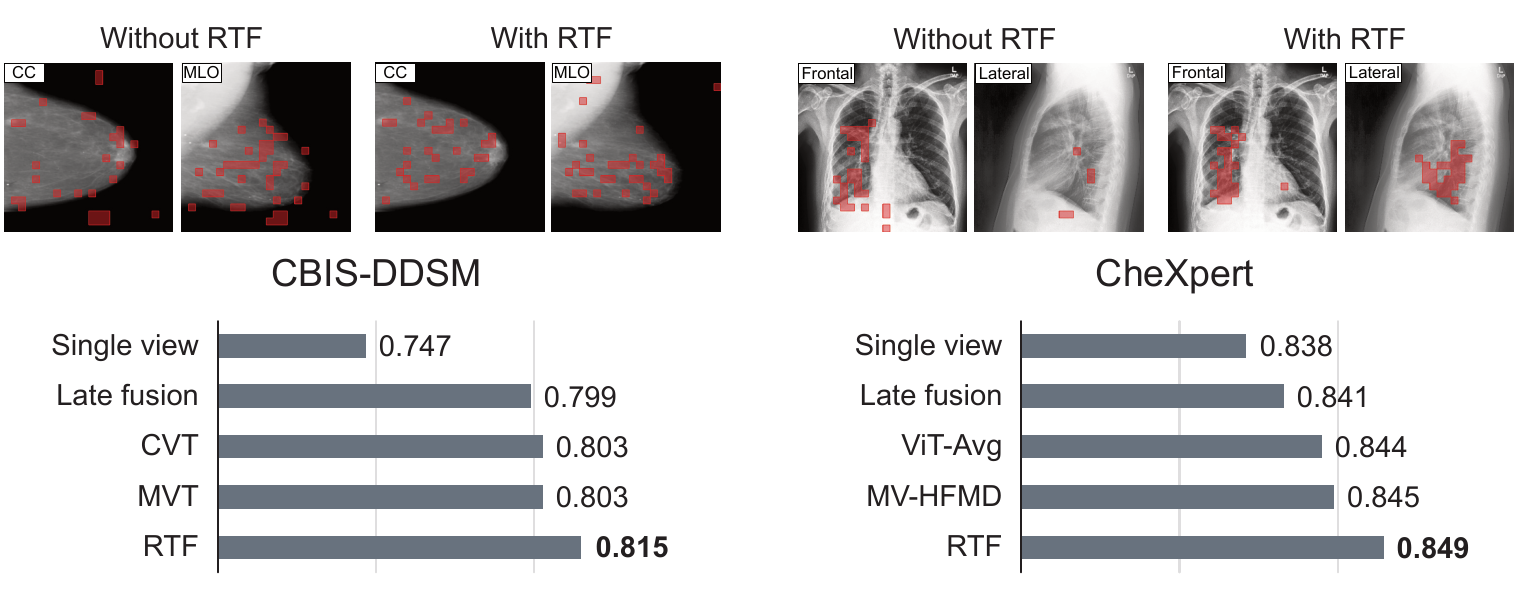}
    \caption{\emph{Illustration of the overfitting problem in multi-view medical diagnosis.} 
    The model's attention becomes overly focused on one of the two available views, resulting in an incomplete interpretation of the case.
    In this example (\textbf{top}), model attention in the MLO view dominates over the CC view in CBIS-DDSM (\textbf{top left}), and the frontal view over the lateral view in CheXpert (\textbf{top right}).
    Random Token Fusion (RTF) encourages the model to better utilize information from both views, resulting in balanced attention between both views and increased performance (\textbf{bottom}).}
    \label{fig:introduction}
\end{figure}

The question of how and where to mix information from different views is an open area of research, with multiple strategies having been proposed in recent years.
This is not our main focus.
Rather, we propose a solution \emph{that can enhance existing multi-view fusion strategies}, which we call Random Token Fusion (RTF).
The motivation for RTF is the observation that models often overfit to the more informative or dominant view, neglecting complementary information in other views
\cite{Wu2020ImprovingTA, wang2020makes}, as illustrated in Figure~\ref{fig:introduction}.
This is particularly problematic in medical image analysis, where each view can provide unique diagnostic insights \cite{feigin_2010, pmlr-v121-hashir20a, raoof_feigin_sung_raoof_irugulpati_rosenow_2012, Irvin_2019}.
Integrating information occlusion and mixing into the training process has been a proven method to combat overfitting and improve robustness \cite{devries2017improved, liu2023patchdropout, Zhang2017mixupBE, yun2019cutmix}.
RTF randomly fuses tokens from different views, introducing variability in the fused representation, which acts as a regularizer. 
This forces the model to consider diverse combinations of tokens from both views, and reduces overfitting to any single view.

RTF can be seamlessly integrated with existing multi-view fusion strategies for vision transformers (ViTs), enriching an existing model's feature space without requiring any modification to the inference process. 
Through extensive experiments on two public medical benchmark datasets, CBIS-DDSM \cite{lee_gimenez_hoogi_miyake_gorovoy_rubin_2017} and CheXpert \cite{Irvin_2019}, we demonstrate that RTF consistently improves performance of multi-view models, achieving state-of-the-art performance on both mammograms and chest X-rays.
The source code used in this work can be found at \href{https://jimyug.github.io/RandomTokenFusion}{https://jimyug.github.io/RandomTokenFusion}. 

\section{Related Work} 
\label{related}

\subsection{Multi-view Fusion in Medical Imaging}
Multi-view fusion is a technique for integrating multiple input signals that typically represent the same object or class 
\cite{yan2021deep}.
It has been used in 3D scene comprehension in the natural domain 
\cite{su2015multi, qi2016volumetric, feichtenhofer2016convolutional, gadzicki2020early}, 
and has recently gained attention in the medical field for improving diagnostics by exploiting complementary information from multiple views taken from the same exam \cite{mane2020multi, Wu2020ImprovingTA}.

For chest X-rays 
\cite{pmlr-v121-hashir20a, 10.1007/978-3-030-87199-4_10}
and mammography screenings 
\cite{WANG201842, 8822935, 8897609, Wu2020ImprovingTA, Chen2022TransformersIB}, studies have demonstrated the utility of merging different views.
Although it has been established that registration is not necessary for fusing multiple views \cite{carneiro2017deep, Chen2022TransformersIB}, simply combining modalities does not guarantee enhanced performance \cite{pmlr-v121-hashir20a, Wu2020ImprovingTA}. 
It is observed that in some cases, one view may overshadow the other in terms of relevance, dominating the learning signal \cite{pmlr-v121-hashir20a, Wu2020ImprovingTA}.
As such, various approaches to multi-view fusion have been investigated.

Wang \etal \cite{WANG201842} adopt an attention mechanism coupled with a recurrent network to merge the two views.
Similarly, Iftikhar \etal \cite{Iftikhar2020MultiViewAL} propose a sensitivity-based weighting mechanism to fuse the predictions on individual views.
Lopez \etal \cite{Lopez2022MultiViewBC} treat the two views as separate channels, constructing a channel-wise input, and employ convolutions to learn the correlations between them.
Liu \etal \cite{liu2021act} employ graph convolutional networks to discern geometric and semantic relationships between two mammography views before merging them. 
MommiNet-v2 \cite{YANG2021102204} emulates radiologists’ practice by integrating symmetry and geometry constraints for improved performance. 
Wu \etal \cite{Wu2020ImprovingTA} randomly omit one of the views during training, forcing the network to utilize both views for improved performance. 
With the emerging trend of vision transformers, their application in medical imaging has great potential with the ability to model complex dependencies across diverse perspectives of subjects. 
Van Tulder \etal \cite{10.1007/978-3-030-87199-4_10} introduce cross-view attention at feature level to transfer information between unregistered views, an approach further proven effective by \cite{10.1007/978-3-031-43904-9_75} 
Moreover, Chen \etal \cite{Chen2022TransformersIB} demonstrate that vision transformers can effectively analyze unregistered multi-view medical images. 
Kim \cite{kim2023chexfusion} and Black \etal \cite{black2024multi} also show that the self-attention mechanism can help aggregate multi-view features efficiently. 

\subsection{Information Mixing as Regularization}
Information occlusion and mixing have been used as a regularization technique in various other contexts. 
Techniques such as CutOut \cite{devries2017improved} and PatchDropout \cite{liu2023patchdropout} have shown their efficacy as regularizers by denying partial information.
Methods such as \cite{Zhang2017mixupBE, yun2019cutmix, faramarzi2022patchup} have demonstrated useful regularization effects when mixing information of different inputs.
Mixup \cite{Zhang2017mixupBE} and CutMix \cite{yun2019cutmix} mix samples at image level during training, while PatchUp \cite{faramarzi2022patchup} works in the hidden space of CNNs.
For ViTs, Liu \etal \cite{Liu2022TokenMixRI} adapt Mixup by blending two images at the token level, guided by contextual activation maps.
Similarly, TokenMixup \cite{2022arXiv221007562C} employs self-attention-based saliency to guide the mixing process towards the most significant tokens.
Zhao \etal \cite{zhao2023mixpro} further advance this approach by optimizing both token and label spaces for ViTs. 

The aforementioned methods provide effective strategies for increasing the diversity of data. 
They have been applied to random image pairs of different objects and labels, although the pairs are intentionally constructed and not intended to be mixed for the specific tasks, \eg. classification. 
Naturally, we believe that the idea could benefit multi-view medical diagnosis where image pairs exist naturally:
They originate from the same object and serve the exact same diagnostic goals, and we want to encourage models to fully utilize both inputs.
To this end, the proposed RTF randomly selects features from both views before fusion, effectively functioning as a regularizer.
By augmenting the feature space, RTF encourages models to analyze cases more comprehensively, leading to improved performance. 

\section{Methods}

\begin{figure}[t]
\centering
\includegraphics[width=\textwidth]{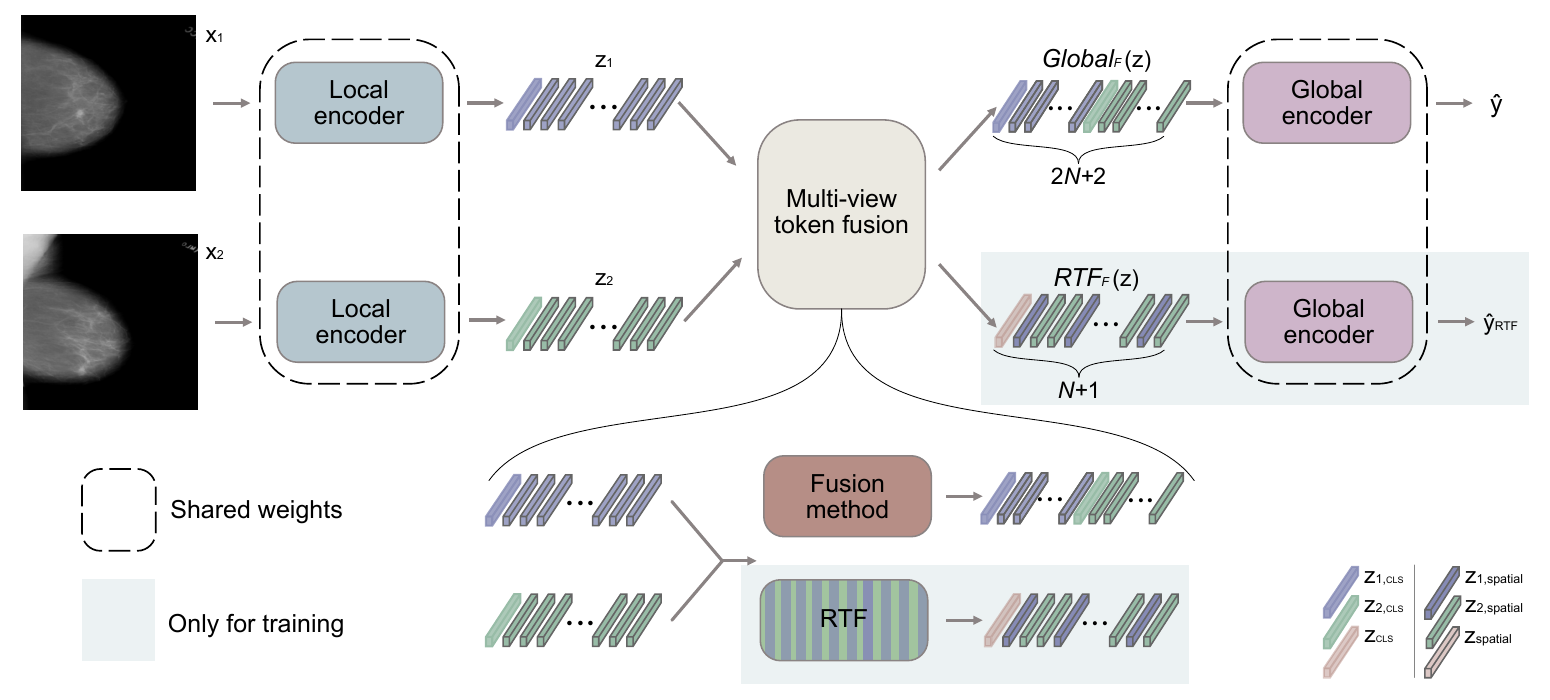}
\caption{
\emph{Multi-view ViTs with Random Token Fusion (RTF).}
RTF utilizes a local encoder to generate representations of different views, followed by a token fusion module. 
This module divides the feature fusion into two distinct branches. 
One branch uses some strategy to merge all tokens from both views, while the other one randomly drops spatial tokens from each view before mixing them. 
The fused tokens are processed by a global encoder, which produces two types of predictions: one for the global tokens and one for the RTF tokens. 
During training, the loss for both branches is minimized towards the same task.
After training, RTF tokens are not generated, they are merged using the model's fusion method and passed to the global encoder for inference.}
\label{fig:framework}
\end{figure}

Combining information from multiple views in medical image analysis enhances diagnostic accuracy, but existing methods often overfit by relying too much on the most informative view \cite{Wu2020ImprovingTA, wang2020makes}. 
To address this, we introduce Random Token Fusion (RTF).
RTF is designed to augment feature learning in multi-view vision transformers by randomly selecting tokens from both views during the fusion phase of training, forcing the model to utilize information from both views effectively (See Figure \ref{fig:framework}).

Given two views \( {x}_{1,2} \in \mathbb{R}^{H \times W \times C} \), where \(H \times W \) is the spatial dimension and \(C\) the number of image channels, a local encoder $f_{\theta_{local}}$ processes each input independently, and generates representations for both views:
\begin{equation} 
    \label{eq:local_module}
    {z}_{1,2} = f_{\theta_{local}}( {x}_{1,2})
\end{equation}
where ${z}_{1,2} \in \mathbb{R}^{\left(N + 1\right) \times D}$, are the local representations of ${x}_{1}$ and ${x}_{2}$, consisting of $N$ patch tokens and a class (CLS) token \cite{Dosovitskiy2020AnII} of size $D$.
Then, ${z}_{1,2}$ are processed by the fusion module that divides them into two different fusion branches: the random token fusion module $RTF_{F}$ (described in \ref{RTF}) and the global fusion module $Global_{F}$ (described in \ref{fusion_strategies}).

Subsequently, the fused outputs are independently forwarded to a global encoder $f_{\theta_{global}}$ that produces the final predictions for both the global and RTF tokens.
The overall process for the two branches is described as follows:
\begin{equation}
    \label{eq:global_branch}
    {\hat{y}} =f_{\theta_{global}}( Global_{F}(f_{\theta_{local}}( {x}_{1,2})))
\end{equation}
for the global branch, and
\begin{equation}
\label{eq:rtf_branch}
    {\hat{y}}_{RTF} =f_{\theta_{global}}( RTF_{F}(f_{\theta_{local}}( {x}_{1,2})))
\end{equation}
for the RTF branch, where ${\hat{y}}$ and ${\hat{y}}_{RTF}$ represent the model's predictions for the global and RTF branch, respectively. 
In this work, we construct the local and the global encoder from the original ViT variants \cite{Dosovitskiy2020AnII} by allocating a certain proportion of the initial \textit{Encoder} blocks to form the local encoder, while the remaining blocks constitute the global encoder.

During training the objective is to minimize a combined loss function, considering both the global and RTF branch
\begin{equation} 
\label{eq:loss}
    \mathcal{L} = \ell_{CE} \left( {\hat{y}}, {y} \right) +  \ell_{CE} \left( {\hat{y}}_{RTF}, {y} \right)
\end{equation}
where $\ell_{CE}$ is the cross-entropy loss between the model's predictions and the target ${y}$.
At inference time, only the process described by Equation \ref{eq:global_branch} is used.

\subsection{Random Token Fusion} 
\label{RTF}
To better instruct the model to utilize information from both views, we introduce the Random Token Fusion (RTF). 
RTF maximizes the mutual information $I(Z; Y)$ between the fused representation \(Z\) and the target \(Y\), mathematically defined as:
\begin{equation}
    I(Z; Y) = H(Z) - H(Z | Y)
\end{equation}
where \(H(Z)\) is the entropy of the fused representation \(Z\), and \(H(Z | Y)\) is the conditional entropy given the target \(Y\).
By incorporating randomness into the token fusion process, RTF encourages the model to learn robust and generalized features from all views, ensuring that the fused representation captures the most informative features.

\begin{figure}[tp]
\centering
\includegraphics[width=\linewidth]{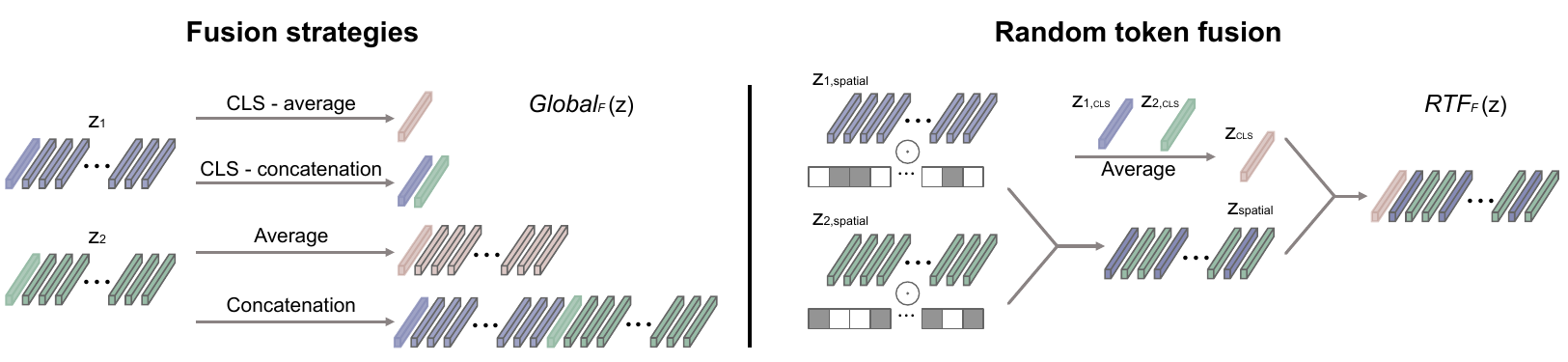}
\caption{
\emph{Illustration of different fusion strategies.}
\textbf{(Left)} Common fusion strategies to fuse the features (tokens) of different views in ViTs.
\textbf{(Right)} The proposed random token fusion (RTF) strategy. In RTF, we randomly drop spatial tokens from both images and combine the remaining ones, augmenting the representations during training.}
\label{fig:RTF}
\end{figure}

RTF randomly selects part of the tokens from each view prior to fusion (see Figure \ref{fig:RTF} Right). 
Specifically, given two features ${z}_{1,2} \in \mathbb{R}^{\left(N+1\right) \times D}$, RTF randomly selects tokens from either view to form a mixed representation 
\begin{equation}
    {z}_{\text{spatial}} = \mathbbm{1}_M \odot {z}_{1,\text{spatial}} + (\mathbbm{1} - \mathbbm{1}_M) \odot {z}_{2,\text{spatial}}, 
\end{equation}
for the spatial tokens, where $\mathbbm{1}_M \in \mathbb{R}^{N}$ is a binary mask, whose elements take the value of $1$ with a probability that follows a uniform distribution \( p \sim \mathcal{U}\left(0,1\right) \).
To preserve high-level features used for classification from both views, we average the CLS tokens $z_{\text{cls}} = \frac{z_{1,\text{cls}} + z_{2,\text{cls}}}{2}$.
We concatenate $z_{\text{cls}}$ and ${z}_{\text{spatial}}$ to form the final fused feature ${z} = \{z_\text{cls}, {z}_{\text{spatial}}\}$.
By doing so, it compels the network to capture dependencies between patches originating from different views, preventing the model from overfitting to view-specific features.

RTF increases both \(H(Z)\) and \(H(Z | Y)\) by introducing randomness in the token selection process. 
Despite the increase in \(H(Z | Y)\), the overall mutual information \(I(Z; Y)\) still benefits because the added randomness enhances the diversity and robustness of the representations, helping the model capture informative complementary features from other views.

\subsection{Fusion Strategies}
\label{fusion_strategies}

RTF is designed to enhance common fusion strategies without bringing additional cost to inference.
There are a number of ways to mix representations of different views.
In this work, we consider the fusion strategies from the literature.

Given the representations of two views ${z}_1$ and ${z}_2$, 
where ${z} = \{z_{\text{cls}}, z_{\text{spatial}_1}, \ldots, z_{\text{spatial}_N}\}$, and $z_{\text{cls}}$ represents the class token, $z_{\text{spatial}_i}$ represents the $i$-th spatial token, and $i$ ranges from $1$ to $N$, 
one strategy is to perform some operation on the features to produce a fused feature ${z}$.
The operation could be token-wise, such as concatenation or selection, or element-wise, such as addition, subtraction, multiplication, max, etc. 

In this work, we consider token-wise fusion with concatenation, a choice widely adopted in current research \cite{Chen2022TransformersIB, 10.1007/978-3-030-87199-4_10}, implemented for ViTs by concatenating all tokens from two views. As such, ${z}$ is defined as 
${z} = \{z_{1,\text{cls}}, \allowbreak z_{1,\text{spatial}_1}, \allowbreak \ldots, z_{1,\text{spatial}_n}, \allowbreak 
z_{2,\text{cls}}, \allowbreak z_{2,\text{spatial}_1}, \ldots, z_{2,\text{spatial}_m}\}$, as depicted in Figure \ref{fig:RTF}.
Since ViTs are able to process an arbitrary number of tokens \cite{10.1007/978-3-030-87199-4_10, Chen2022TransformersIB}, this is a trivial operation.
While this approach has the benefit of preserving information, it adds a considerable cost to the memory and compute. 
A compromised solution would be keeping and fusing only the CLS tokens 
$z = \{z_{1_{\text{cls}}} , z_{2_{\text{cls}}}\}$ 
for efficiency.

An alternative approach is element-wise fusion by averaging \cite{Wu2020ImprovingTA}, implemented by averaging all tokens from two views ${z} = \frac{{z}_{1} + {z}_{2}}{2}$ or by using only the CLS tokens, $z = \frac{z_{1,cls} + z_{2,cls}}{2}$ (see Figure \ref{fig:RTF}). 
While this approach is straightforward and computationally efficient, it may lead to loss of spatial information, particularly when input images are not registered \cite{10.1007/978-3-030-87199-4_10}.

\section{Experimental Setup} 
\label{experiments}

By design, RTF can be applied to most transformer-based multi-view fusion models.
In this work, we employ the standard ViT family \cite{Dosovitskiy2020AnII} for both the local and global encoders.
Using ViT Tiny, Small, and Base, we conduct experiments on two standard benchmark medical datasets, CBIS-DDSM \cite{lee_gimenez_hoogi_miyake_gorovoy_rubin_2017, CBIS_DDSM_Citation} and CheXpert \cite{Irvin_2019} to evaluate RTF performance on top of different fusion strategies, described in Section \ref{fusion_strategies}.
We also compare against other multi-view fusion methods designed for ViTs and, for reference, against CNN-based multi-view fusion methods.
For both datasets, we use the area under the receiver operating characteristic curve (AUC) to evaluate model performance.
All experiments are repeated $4$ times, and we report the mean value and standard deviation.

\begin{table}[t]
\caption{The effect of using only a single view, multiple views with late fusion, and multiple views with RTF on CBIS-DDSM (\textbf{left}) and CheXpert (\textbf{right}).} 
\centering
\begin{tabular}{l@{\hspace{5mm}}cl@{\hspace{12mm}}lc}
\cmidrule[0.8pt](l{-0.1mm}r{0.1mm}){0-1}
\cmidrule[0.8pt](l{-2.5mm}r{0.1mm}){4-5}

\textbf{View} &
\textbf{DDSM}, AUC $\uparrow$ &
&
\textbf{View} &
\textbf{CheXpert}, AUC $\uparrow$ 
\\ 
\cmidrule[0.45pt](l{-0.1mm}r{0.1mm}){0-1}
\cmidrule[0.45pt](l{-2.5mm}r{0.1mm}){4-5}
{Only CC} &
0.730 $\pm$ 0.004 &
&
Only Frontal &
0.838 $\pm$ 0.003
\\
{Only MLO} &
0.747 $\pm$ 0.022 &
&
Only Lateral &
0.832 $\pm$ 0.002
\\
\cmidrule[0.45pt](l{-0.1mm}r{0.1mm}){0-1}
\cmidrule[0.45pt](l{-2.5mm}r{0.1mm}){4-5}
{Late fusion} &
0.799 $\pm$ 0.008 &
&
{Late fusion} &
0.841 $\pm$ 0.001
\\
{Fusion w/ RTF} &
\textbf{0.815 $\pm$ 0.001} &
&
{Fusion w/ RTF} &
\textbf{0.849 $\pm$ 0.001}
\\
\cmidrule[0.8pt](l{-0.1mm}r{0.1mm}){0-1}
\cmidrule[0.8pt](l{-2.5mm}r{0.1mm}){4-5}

\end{tabular} 
  
\label{tab:single_multi_rtf}
\end{table}

\subsection{Datasets}
\paragraph{CBIS-DDSM \cite{lee_gimenez_hoogi_miyake_gorovoy_rubin_2017, CBIS_DDSM_Citation}} is a well-known public mammography dataset, containing $10,239$ samples from $1,566$ unique patients, including both craniocaudal (CC) and mediolateral oblique (MLO) views.
The objective is to classify mass abnormalities as either benign or malignant.
We only select images from exams for which both CC and MLO views are available, similar to \cite{10.1007/978-3-030-87199-4_10}.
This results in a subset of $1,416$ images from $708$ exams, corresponding to $636$ patients.

\paragraph{CheXpert \cite{Irvin_2019}} is one of the largest publicly available datasets for chest radiograph interpretation, consisting of frontal and lateral chest X-ray scans. 
The original dataset contains $224,316$ chest radiographs of $65,240$ patients, annotated for $13$ diagnostic observations, each labelled as negative, positive, uncertain, or unknown (missing).
The uncertain and unknown samples are handled following the suggestion of \cite{Irvin_2019}.
For the purposes of this work, we only use exams that include both frontal and lateral views, totaling $62,826$ images from $31,413$ exams ($22,414$ patients).

\subsection{Implementation Details}
We resize all images to $384 \times 384$ and initialize all models with weights pretrained on ImageNet-21K \cite{ILSVRC15, rw2019timm}, as randomly initialized ViTs cannot effectively be trained on small medical datasets \cite{matsoukas2021time}.
By default, we use ViT Small \cite{Dosovitskiy2020AnII} as the backbone, with concatenation as the fusion strategy.
$75\%$ of the total encoder blocks are used as the local encoder $f_{\theta_{local}}$, and the remaining $25\%$ are applied to the fused feature.
This partitioning is used by default throughout the rest of our experiments.
We use the AdamW optimizer \cite{Loshchilov2017DecoupledWD} to train the models for $300$ epochs on CBIS-DDSM and for $60$ epochs on CheXpert. 
The learning rate is selected based on a grid search.
For both datasets, we use spatial scaling, flipping, rotation, and color jittering for data augmentation. 
After training, the checkpoint with the highest AUC validation score is selected for testing. 
We employ the same train/validation/test split as in \cite{10.1007/978-3-030-87199-4_10, black2024multi} for both datasets.
All experiments are conducted on NVIDIA GeForce RTX 4090 Ti GPUs with 24 GB of memory.

\addtolength{\tabcolsep}{+4.2pt}
\begin{table}[b]
\caption{Performance (AUC) depending on where the encoder is split for fusion, w/{\scriptsize(w/o)} RTF.}
\centering
\begin{adjustbox}{max width=\linewidth}
\begin{tabular}{lccc}
\toprule
\textbf{Local/ Global} &
\textbf{25\%/ 75\%} &
\textbf{50\%/ 50\%} &
\textbf{75\%/ 25\%} 
\\ 
\midrule
CBIS-DDSM &
0.802 {\scriptsize(0.799)} &
0.810 {\scriptsize(0.799)} &
0.815 {\scriptsize(0.803)}
\\
CheXpert &
0.845 {\scriptsize(0.842)} &
0.849 {\scriptsize(0.844)} &
0.849 {\scriptsize(0.843)} 
\\
\bottomrule

\end{tabular} 
 
\end{adjustbox}
\label{tab:local_global_ratio}
\end{table}
\addtolength{\tabcolsep}{-4.2pt}

\section{Results and Discussion}
\label{results}

\subsection{Ablation Study}
\label{text:ablation}

We begin by confirming the advantages of multi-view over single-view ViTs across various fusion strategies.
We then show the benefits of incorporating RTF with standard fusion techniques.
We then delve into identifying the most compatible fusion strategy for RTF, and conduct ablation studies for different model sizes.

\paragraph{Are there benefits to using multiple views compared to a single view?}
To assess the benefits of multi-view ViTs, we conducted experiments training models on single views and with late fusions and report results in Table \ref{tab:single_multi_rtf}.
Specifically, single-view models only consider one view throughout the task, while late-fusion models combine representations of different views at the end of the model \cite{carneiro2017deep, black2024multi, 10.1007/978-3-030-87199-4_10}. 
We verify that, for both CBIS-DDSM and CheXpert, using two views results in improved AUC performance compared to single-view models with the same capacity, and fusing information at feature level is more effective than at a late stage, which is consistent with prior findings \cite{pmlr-v121-hashir20a, Wu2020ImprovingTA}.
This also highlights the importance of encouraging models to utilize both perspectives in medical diagnosis, which is a key motivation for our research.

To determine the optimal point for fusing local features, we conducted experiments to assess its impact on performance. 
We derived two key observations from the results \footnote{Extended tables can be found in the Supplement.} in Table \ref{tab:local_global_ratio}.
First, the advantage of integrating multiple views within a model, as opposed to late fusion, generally persists. 
On both datasets, effective intermediate feature fusion outperforms late fusion, as shown in Tables \ref{tab:single_multi_rtf} and  \ref{tab:local_global_ratio}.
Second, RTF consistently enhances performance across all settings. 
The best results are achieved when the first $75\%$ of the blocks are used for the local encoder, and the remaining $25\%$ blocks for the global encoder.
We adopt this as the default setting in other experiments.

\paragraph{What are the benefits from using RTF to enhance multi-view fusion?}
To understand the impact of integrating RTF with common fusion strategies, we train models of different capacities employing the three token fusion strategies described in Section \ref{fusion_strategies}: averaging all tokens (Average), concatenating only the CLS tokens (CLS$_{\text{cat}}$), and concatenating all tokens (Concat) from two views, with and without RTF. We present the results on CBIS-DDSM and CheXpert in Table \ref{tab:ablation-ddsm} and Table \ref{tab:ablation-chexpert}, respectively.

The results show that training with RTF consistently improves performance across all configurations.
And the extent of improvement varies with the dataset and model size. 
CBIS-DDSM appears to gain more from RTF, particularly for larger ViT variants. 
We hypothesize that this is due to the regularization effects of RTF and the smaller size of the dataset, as higher-capacity models are more prone to overfitting \cite{hinton2012improving}.
Overall, integrating RTF with concatenation fusion for the global encoder generally yields better performance compared to other fusion methods, as it preserves the tokens and retains the most information.
Therefore, we adopt this combination as the default setting for later comparisons.

\addtolength{\tabcolsep}{+5.0pt}
\begin{table}[tp]
\centering
\caption{
AUC performance on CBIS-DDSM, showing the effect of using multiple views with and without RTF for different model sizes and fusion strategies.
}
\begin{adjustbox}{max width=\linewidth}
\begin{tabular}{lcccc}
\toprule
\textbf{Method} &
\textbf{RTF Used} &
\textbf{ViT Tiny} &
\textbf{ViT Small} &
\textbf{ViT Base}
\\ 
\midrule
\multirow{2}{*}{Average} &
No & 0.798 $\pm$ 0.003 &
0.803 $\pm$ 0.008 &
0.813 $\pm$ 0.004
\\
& {Yes} &
\textbf{0.802 $\pm$ 0.001} &
\textbf{0.809 $\pm$ 0.002} &
\textbf{0.825 $\pm$ 0.005}
\\
\midrule
\multirow{2}{*}{CLS$_{\text{cat}}$} &
No &
0.796 $\pm$ 0.002 &
0.802 $\pm$ 0.006 &
0.814 $\pm$ 0.007
\\
& {Yes} &
\textbf{0.801 $\pm$ 0.001} &
\textbf{0.811 $\pm$ 0.008} &
\textbf{0.826 $\pm$ 0.004}
\\
\midrule
\multirow{2}{*}{Concat} &
No &
0.798 $\pm$ 0.003 &
0.803 $\pm$ 0.003 &
0.814 $\pm$ 0.004
\\
& {Yes} &
\textbf{0.802 $\pm$ 0.003} &
\textbf{0.815 $\pm$ 0.001} &
\textbf{0.830 $\pm$ 0.002}
\\
\bottomrule

\end{tabular} 
 
\end{adjustbox}
\label{tab:ablation-ddsm}
\end{table}
\addtolength{\tabcolsep}{-5.0pt}

\addtolength{\tabcolsep}{+5.0pt}
\begin{table}[tp]
\centering
\caption{AUC performance on CheXpert, showing the effect of using multiple views with and without RTF for different model sizes and fusion strategies.}
\begin{adjustbox}{max width=\linewidth}
\begin{tabular}{lcccc}
\toprule
\textbf{Method} &
\textbf{RTF Used} &
\textbf{ViT Tiny} &
\textbf{ViT Small} &
\textbf{ViT Base}
\\ 
\midrule
\multirow{2}{*}{Average} &
No & 
0.835 $\pm$ 0.003 &
0.844 $\pm$ 0.004 &
0.847 $\pm$ 0.003
\\
& {Yes} &
\textbf{0.838 $\pm$ 0.003} &
\textbf{0.848 $\pm$ 0.002} &
\textbf{0.850 $\pm$ 0.002}
\\
\midrule
\multirow{2}{*}{CLS$_{\text{cat}}$} &
No & 
0.833 $\pm$ 0.003 &
0.842 $\pm$ 0.003 &
0.846 $\pm$ 0.002
\\
& {Yes} &
\textbf{0.837 $\pm$ 0.003} &
\textbf{0.846 $\pm$ 0.001} &
\textbf{0.850 $\pm$ 0.001}
\\
\midrule
\multirow{2}{*}{Concat} &
No & 
0.833 $\pm$ 0.002 &
0.843 $\pm$ 0.004 &
0.847 $\pm$ 0.004
\\
& {Yes} &
\textbf{0.839 $\pm$ 0.002} &
\textbf{0.849 $\pm$ 0.001} &
\textbf{0.852 $\pm$ 0.002}
\\
\bottomrule
\end{tabular}
 
\end{adjustbox}
\label{tab:ablation-chexpert}
\end{table}
\addtolength{\tabcolsep}{-5.0pt}

Inspired by recent advancements in hybrid architectures \cite{kim2023chexfusion, black2024multi}, we aim to examine the efficacy of RTF with different backbone choices. 
Compared to the plain ViT \cite{Dosovitskiy2020AnII}, hybrid architectures adopt more complex and powerful designs, \eg, the ResNet family or other CNNs, to project input images to tokens before feeding them into a transformer-based network.
Following \cite{black2024multi}, we employ three variants of the ``ResNet + ViT'' design and apply RTF between the ResNet and ViT Encoder. 
We apply RTF to the tokens generated by the local CNN encoders and follow the same procedure outlined in Figure \ref{fig:framework}. As shown in Table \ref{tab:ablation-hybrid}, RTF consistently benefits the hybrid models.
Although hybrid architectures generally achieve better performance than plain ViTs, they are computationally expensive and require larger memory \cite{Dosovitskiy2020AnII, rw2019timm}. 
As such, we opt for standard ViTs as the main backbone in this work.

\subsection{Comparison with State of the Art}
We benchmark RTF against state-of-the-art (SOTA) methods from the literature.
On CBIS-DDSM, we include ResNet50 \cite{Lopez2022MultiViewBC}, Shared ResNet \cite{Wu2020ImprovingTA}, PHResNet50 \cite{Lopez2022MultiViewBC}, Multi-view Transformer (MVT) \cite{Chen2021MVT, Chen2022TransformersIB}, ViT with average fusion (ViT-Average) \cite{nguyen2022novel}, and Cross-view Transformer (CVT) \cite{10.1007/978-3-030-87199-4_10}. 
On CheXpert, we compare RTF with 
MVC-NET \cite{Zhu2021MVC}, MVCNN \cite{Su2015MultiviewCN}, MVT \cite{Chen2021MVT, Chen2022TransformersIB}, ViT-Average \cite{nguyen2022novel}, CVT \cite{10.1007/978-3-030-87199-4_10}, and MV-HFMD \cite{black2024multi}.
We choose RTF Small, \ie, ViT Small with RTF, which has a comparable capacity with other models. 

As shown in Table \ref{tab:comparison}, RTF outperforms all methods on both datasets, showing its efficacy in multi-view medical diagnosis. 
Notably, RTF can be used in conjunction with transformer-based methods, such as ViT-Average \cite{nguyen2022novel} and MVT \cite{Chen2021MVT, Chen2022TransformersIB}, for further enhanced performance, as demonstrated in Section \ref{text:ablation}.
By design, RTF during inference is essentially a standard ViT \cite{Dosovitskiy2020AnII} with concatenation for feature fusion. 
The fact that RTF achieves better performance on CheXpert than MV-HFMD \cite{black2024multi} -- a hybrid architecture -- highlights the potential of ViTs in this application when empowered with RTF.
It is important to note that comparing RTF against CNN-based methods may place the latter at a disadvantage, as evidence suggests that transformers are more adept at fusion tasks \cite{Chen2022TransformersIB, al-hammuri_gebali_kanan_chelvan_2023}. 
Nevertheless, these comparisons are presented for completeness.

\begin{table}[t]
\caption{Comparison vs. SOTA methods on CBIS-DDSM (\textbf{left}) and CheXpert (\textbf{right}).} 
\centering
\begin{tabular}{lc@{\hspace{1mm}}c@{\hspace{12mm}}lc}
\cmidrule[0.8pt](l{-0.1mm}r{0.1mm}){1-2}
\cmidrule[0.8pt](l{-0.1mm}r{0.1mm}){4-5}
\textbf{Method} &
\textbf{CBIS-DDSM} &
&
\textbf{Method} &
\textbf{CheXpert}
\\
\cmidrule[0.45pt](l{-0.1mm}r{0.1mm}){1-2}
\cmidrule[0.45pt](l{-0.1mm}r{0.1mm}){4-5}
ResNet50 \cite{Lopez2022MultiViewBC}  & 0.724 $\pm$ 0.007
&
&
MVC-NET \cite{Zhu2021MVC}                & 0.813 $\pm$ 0.005
\\
Shared ResNet \cite{Wu2020ImprovingTA}   & 0.735 $\pm$ 0.014
&
&
MVCNN \cite{Su2015MultiviewCN}           & 0.815 $\pm$ 0.004
\\
PHResNet50 \cite{Lopez2022MultiViewBC}   & 0.739 $\pm$ 0.004
&
&
CVT
\cite{10.1007/978-3-030-87199-4_10}      & 0.834 $\pm$ 0.002
\\
MVT \cite{Chen2021MVT, Chen2022TransformersIB}         
                                         & 0.803 $\pm$ 0.003
&
&
MVT \cite{Chen2021MVT, Chen2022TransformersIB}            
                                         & 0.843 $\pm$ 0.004
\\
CVT
\cite{10.1007/978-3-030-87199-4_10}      & 0.803 $\pm$ 0.007
&
&
ViT-Average \cite{nguyen2022novel}       & 0.844 $\pm$ 0.004
\\
ViT-Average \cite{nguyen2022novel}       & 0.803 $\pm$ 0.008
&
&
MV-HFMD \cite{black2024multi}            & 0.845 $\pm$ 0.002
\\
\textbf{RTF}    & \textbf{0.815 $\pm$ 0.001}
&
&
\textbf{RTF}    & \textbf{0.849 $\pm$ 0.001}
\\
\cmidrule[0.8pt](l{-0.1mm}r{0.1mm}){1-2}
\cmidrule[0.8pt](l{-0.1mm}r{0.1mm}){4-5}
\end{tabular}
  
\label{tab:comparison}
\end{table}

\begin{table}[b]
\caption{Performance (AUC) on CheXpert of hybrid models, w/ and w/o RTF.} 
\centering
\begin{tabular}{lccc}
\toprule
\textbf{Fusion} & 
\textbf{R+ViT-Ti/16} & 
\textbf{R26+ViT-S/16} &
\textbf{R50+ViT-B/16}
\\ 
\midrule
w/o RTF & 
0.834 $\pm$ 0.002 &
0.846 $\pm$ 0.002 &
0.848 $\pm$ 0.001
\\
w/ RTF & 
0.839 $\pm$ 0.002 &
0.852 $\pm$ 0.002 &
0.854 $\pm$ 0.002
\\ 
\bottomrule
\end{tabular}
  
\label{tab:ablation-hybrid}
\end{table}
\addtolength{\tabcolsep}{+1.3pt}

\subsection{Qualitative Results}
\begin{figure}[t]
\includegraphics[width=\linewidth]{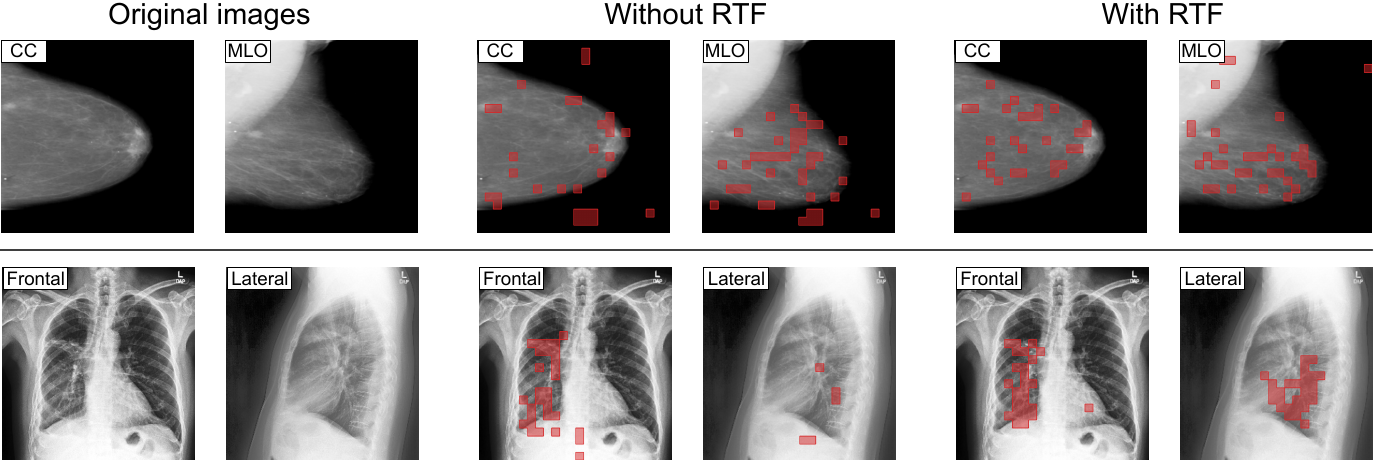}
\caption{
\emph{RTF encourages the model to learn from both views, resulting in better quality of attention maps.}
Given two views (\textbf{left}), standard fusion methods overfit on a specific view (MLO in CBIS-DDSM and Frontal in CheXpert, in this example), under-utilizing information from the other (\textbf{middle}).
RTF encourages the model to use information for both views to make better predictions (\textbf{right}).
For mammograms, the model sometimes focuses on uninformative background, a known issue raised when information of the patches is redundant, and may lead to poor generalization. 
With RTF, the quality of the attention maps is better in general, with more attention focused on relevant areas. 
On CheXpert, the baseline model ignores the lateral view that also provide valuable information for diagnosis. 
RTF also seems to encourage more equal focus between both views. }
\label{fig:attention}
\end{figure}

The intuition behind RTF is that by withholding information during training, the model is encouraged to focus its attention on features from both views. 
To better understand the impact of RTF on model attention, we provide qualitative examples from CBIS-DDSM and CheXpert. 
In Figure \ref{fig:attention}, we compare attention maps within the last block of the global encoder from models trained with and without RTF, with each row showing the results of an image pair with two views.
On CBIS-DDSM, the baseline model sometimes focuses on uninformative background, a phenomenon observed where patch information is redundant, as discussed in \cite{darcet2023vision}. 
This may lead to poor generalization. 
With RTF, the quality of the attention maps generally improves, with more attention focused on relevant areas. 
RTF also appears to encourage a more balanced focus between both views in many cases.
On CheXpert, the model with RTF tends to consider both views, while the baseline model often ignores the lateral view, which may provide valuable information for diagnosis \cite{pmlr-v121-hashir20a, Irvin_2019}.
More examples are included in the Supplement.

\subsection{Further Discussion}
\label{sec:further}

\paragraph{Can RTF work as a standalone, efficient training method?}
In this study, we propose RTF as a solution to enhance existing multi-view fusion strategies. 
RTF is designed to function with an additional branch in the network and loss function, making it compatible with various manipulations of intermediate features (tokens). 
Some approaches corrupt spatial information, while others retain all tokens as-is or completely discard the spatial ones. 
It would be intriguing to explore whether RTF alone could serve as an effective fusion strategy, potentially reducing computational requirements through random token selection compared to simple concatenation, and further improving performance. 
The potential positive outcomes could contribute to both efficient training and robust medical diagnosis. 
We plan to investigate this further in future work.

\section{Conclusion}
\label{sec:concluddion}

In this work, we focus on multi-view vision transformers for medical image analysis.
We address shortcomings of current fusion methods that tend to overfit on view-specific features, not fully leveraging information from all views.
Random token fusion (RTF) randomly drops tokens from both views during the fusion phase of training, encouraging the model to learn more robust representations across all views.
RTF directly impacts the model's attention and enhances its performance without any additional cost at inference.
Our experiments show that RTF exceeds the performance of other fusion methods and seamlessly boosts performance when combined with them -- the degree of improvement is influenced by the dataset and model size.
While our work focuses on multi-view ViTs for medical diagnosis with chest X-rays and mammograms, we believe that our findings can be extended to multiple medical modalities, as well as other tasks and domains, which we leave for future work.

\paragraph{Acknowledgements.} 
This work was partially supported by the Wallenberg AI, Autonomous Systems and Software Program (WASP) funded by the Knut and Alice Wallenberg Foundation. 

\clearpage

% %%%%%%%%% REFERENCES
{
\small
\printbibliography
}

% Appendix
\clearpage
{\center\baselineskip 18pt
    \vskip .25in{\Large\bf
    Appendix \par
}\vskip 0.35in}

\begin{appendices}
\renewcommand{\thesection}{\Alph{section}}%

\section{Extended Results}

We report additional experimental results included in the main paper. 
Table \ref{tab:appendix-ratio-ddsm} and Table \ref{tab:appendix-ratio-che} extend Table 2 of the main paper. 
We investigate how many blocks of a standard ViT are used as the local encoder and the remaining as the global encoder for the fused tokens with all mentioned fusion strategies. 
Two key findings emerge from the results:

First, concatenation proves more robust to the choice of local/global ratio compared to the other fusion strategies. 
This robustness is expected, as concatenation preserves the information to the most extent and can fully utilize the global transformer blocks.
Based on these results, we select concatenation as the default fusion strategy. 
Second, RTF generally enhances performance across all settings. 
The only exception occurs when using $25\%$ blocks as the local encoder with CLS$_{\text{cat}}$.
In this scenario, all spatial tokens are discarded at a very early stage, and only the two CLS tokens are sent to the global encoder, resulting in extremely low model capacity. 
Applying RTF in this situation harms performance, similar to the effects of aggressive regularization techniques on an already under-fitting model.

\addtolength{\tabcolsep}{+5.0pt}
\begin{table}[htbp]
\centering
\caption{
Extended results on CBIS-DDSM, showing AUC performance depending on where the encoder is split for fusion.}
\label{tab:appendix-ratio-ddsm}
\begin{tabular}{lccccc}
\toprule
\textbf{Fusion} & 
\textbf{RTF Used} & 
\textbf{25\% local} &
\textbf{50\% local} &
\textbf{75\% local} \\ 
\midrule
\multirow{2}{*}{Average} & 
No & 
0.753 $\pm$ 0.007 & 
0.789 $\pm$ 0.014 & 
0.803 $\pm$ 0.008 \\
& Yes & 
\textbf{0.756 $\pm$ 0.011} & 
\textbf{0.793 $\pm$ 0.006} & 
\textbf{0.809 $\pm$ 0.002} \\ 
\cmidrule{1-5}
\multirow{2}{*}{CLS$_{\text{cat}}$} & 
No & 
\textbf{0.711 $\pm$ 0.012} & 
0.782 $\pm$ 0.007 & 
0.802 $\pm$ 0.006 \\
& Yes &
0.709 $\pm$ 0.005 & 
\textbf{0.796 $\pm$ 0.001} & 
\textbf{0.811 $\pm$ 0.008} \\ 
\cmidrule{1-5}
\multirow{2}{*}{Concat} & 
No & 
0.799 $\pm$ 0.002 & 
0.799 $\pm$ 0.009 &
0.803 $\pm$ 0.003 \\
& Yes & 
\textbf{0.802 $\pm$ 0.001} & 
\textbf{0.810 $\pm$ 0.003} & 
\textbf{0.815 $\pm$ 0.001} \\ 
\bottomrule
\end{tabular}

\end{table}
\addtolength{\tabcolsep}{-5.0pt}

\addtolength{\tabcolsep}{+5.0pt}
\begin{table}[htbp]
\centering
\caption{
Extended results on CheXpert, showing AUC performance depending on where the encoder is split for fusion.}
\label{tab:appendix-ratio-che}
\begin{adjustbox}{max width=\textwidth}
\begin{tabular}{lccccc}
\toprule
\textbf{Fusion} & 
\textbf{RTF Used} &     
\textbf{25\% local} &
\textbf{50\% local} &
\textbf{75\% local} \\ 
\midrule
\multirow{2}{*}{Average} & 
No & 
0.834 $\pm$ 0.004 & 
0.845 $\pm$ 0.002 & 
0.844 $\pm$ 0.004 \\
& Yes & 
\textbf{0.835 $\pm$ 0.003} & 
\textbf{0.849 $\pm$ 0.001} & 
\textbf{0.848 $\pm$ 0.002} \\ 
\cmidrule{1-5}
\multirow{2}{*}{CLS$_{\text{cat}}$} & 
No & 
\textbf{0.815 $\pm$ 0.003} & 
0.841 $\pm$ 0.003 & 
0.842 $\pm$ 0.006 \\
& Yes &
0.814 $\pm$ 0.003 & 
\textbf{0.844 $\pm$ 0.001} & 
\textbf{0.846 $\pm$ 0.001} \\ 
\cmidrule{1-5}
\multirow{2}{*}{Concat} & 
No & 
0.842 $\pm$ 0.003 & 
0.844 $\pm$ 0.003 &
0.843 $\pm$ 0.004 \\
& Yes & 
\textbf{0.845 $\pm$ 0.002} & 
\textbf{0.849 $\pm$ 0.001} & 
\textbf{0.849 $\pm$ 0.001} \\ 
\bottomrule
\end{tabular}

\end{adjustbox}
\end{table}
\addtolength{\tabcolsep}{-5.0pt}

\newpage
\section{Additional Saliency Results}

\begin{figure}[htbp]
\centering
\includegraphics[width=1.0\linewidth]{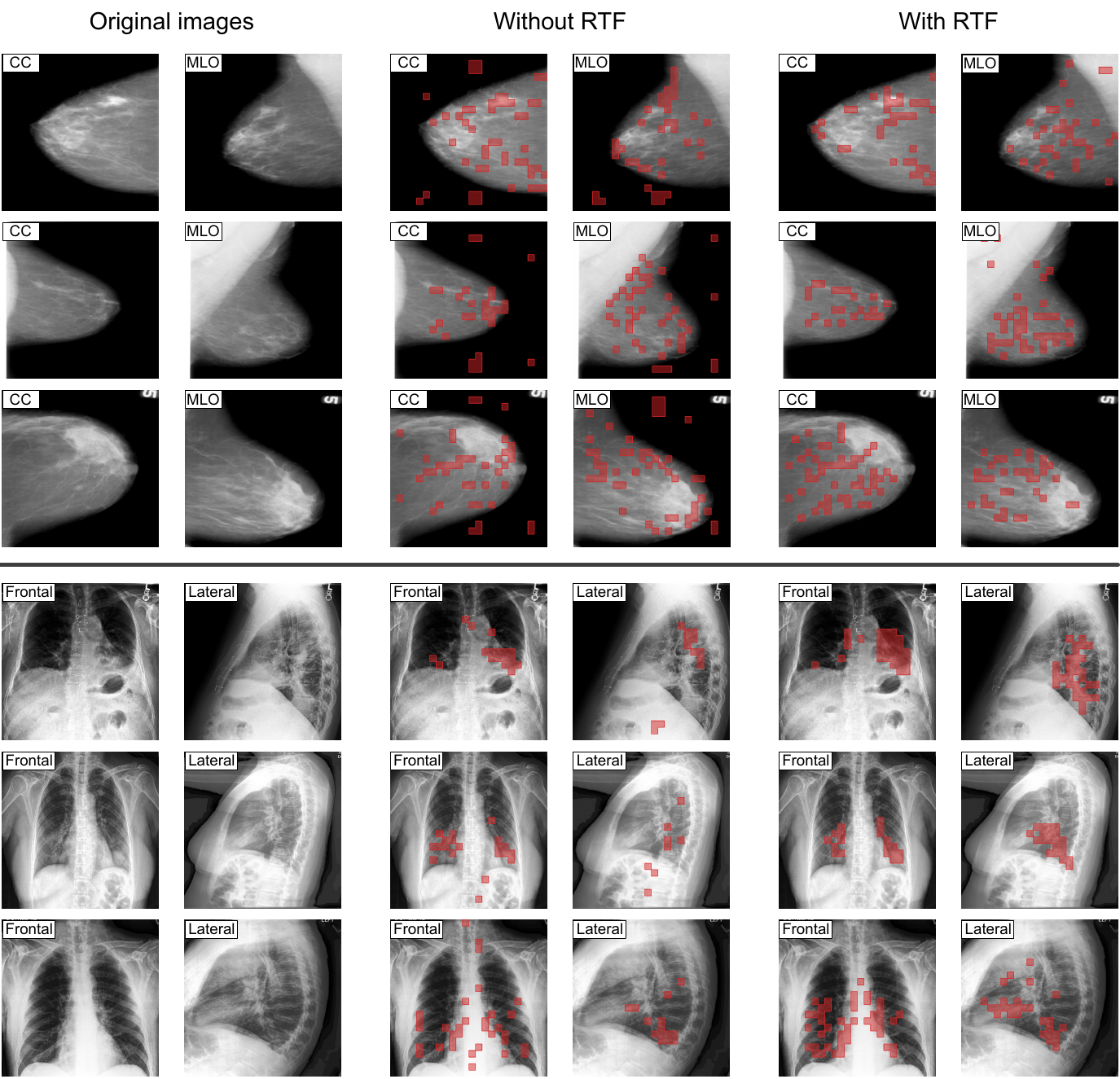}
\caption{
Extended results on CBIS-DDSM (\textbf{top}) and CheXpert (\textbf{bottom}), showing the model's attention maps within the last block of the global encoder.
RTF seems to address the issue of attention being allocated to uninformative areas, a common phenomenon observed in ViTs.
It also encourages the model to focus on both views in many cases.}
\label{fig:attention_appendix}
\end{figure} 

\end{appendices}

% Chekclist
% \clearpage
% \input{8_NeurIPS_Checklist}

\end{document}